\documentclass{acmproc-sp}
\usepackage[draft]{minted}
\usepackage{courier}
\usepackage{listings}

\makeatletter
\def\PYGdefault@reset{\let\PYGdefault@it=\relax \let\PYGdefault@bf=\relax%
    \let\PYGdefault@ul=\relax \let\PYGdefault@tc=\relax%
    \let\PYGdefault@bc=\relax \let\PYGdefault@ff=\relax}
\def\PYGdefault@tok#1{\csname PYGdefault@tok@#1\endcsname}
\def\PYGdefault@toks#1+{\ifx\relax#1\empty\else%
    \PYGdefault@tok{#1}\expandafter\PYGdefault@toks\fi}
\def\PYGdefault@do#1{\PYGdefault@bc{\PYGdefault@tc{\PYGdefault@ul{%
    \PYGdefault@it{\PYGdefault@bf{\PYGdefault@ff{#1}}}}}}}
\def\PYGdefault#1#2{\PYGdefault@reset\PYGdefault@toks#1+\relax+\PYGdefault@do{#2}}

\expandafter\def\csname PYGdefault@tok@k\endcsname{\let\PYGdefault@bf=\textbf\def\PYGdefault@tc##1{\textcolor[rgb]{0.00,0.50,0.00}{##1}}}
\expandafter\def\csname PYGdefault@tok@s2\endcsname{\def\PYGdefault@tc##1{\textcolor[rgb]{0.73,0.13,0.13}{##1}}}
\expandafter\def\csname PYGdefault@tok@c1\endcsname{\let\PYGdefault@it=\textit\def\PYGdefault@tc##1{\textcolor[rgb]{0.25,0.50,0.50}{##1}}}
\expandafter\def\csname PYGdefault@tok@kn\endcsname{\let\PYGdefault@bf=\textbf\def\PYGdefault@tc##1{\textcolor[rgb]{0.00,0.50,0.00}{##1}}}
\expandafter\def\csname PYGdefault@tok@mb\endcsname{\def\PYGdefault@tc##1{\textcolor[rgb]{0.40,0.40,0.40}{##1}}}
\expandafter\def\csname PYGdefault@tok@nl\endcsname{\def\PYGdefault@tc##1{\textcolor[rgb]{0.63,0.63,0.00}{##1}}}
\expandafter\def\csname PYGdefault@tok@s1\endcsname{\def\PYGdefault@tc##1{\textcolor[rgb]{0.73,0.13,0.13}{##1}}}
\expandafter\def\csname PYGdefault@tok@gd\endcsname{\def\PYGdefault@tc##1{\textcolor[rgb]{0.63,0.00,0.00}{##1}}}
\expandafter\def\csname PYGdefault@tok@sb\endcsname{\def\PYGdefault@tc##1{\textcolor[rgb]{0.73,0.13,0.13}{##1}}}
\expandafter\def\csname PYGdefault@tok@mo\endcsname{\def\PYGdefault@tc##1{\textcolor[rgb]{0.40,0.40,0.40}{##1}}}
\expandafter\def\csname PYGdefault@tok@kd\endcsname{\let\PYGdefault@bf=\textbf\def\PYGdefault@tc##1{\textcolor[rgb]{0.00,0.50,0.00}{##1}}}
\expandafter\def\csname PYGdefault@tok@go\endcsname{\def\PYGdefault@tc##1{\textcolor[rgb]{0.53,0.53,0.53}{##1}}}
\expandafter\def\csname PYGdefault@tok@ni\endcsname{\let\PYGdefault@bf=\textbf\def\PYGdefault@tc##1{\textcolor[rgb]{0.60,0.60,0.60}{##1}}}
\expandafter\def\csname PYGdefault@tok@nt\endcsname{\let\PYGdefault@bf=\textbf\def\PYGdefault@tc##1{\textcolor[rgb]{0.00,0.50,0.00}{##1}}}
\expandafter\def\csname PYGdefault@tok@c\endcsname{\let\PYGdefault@it=\textit\def\PYGdefault@tc##1{\textcolor[rgb]{0.25,0.50,0.50}{##1}}}
\expandafter\def\csname PYGdefault@tok@gt\endcsname{\def\PYGdefault@tc##1{\textcolor[rgb]{0.00,0.27,0.87}{##1}}}
\expandafter\def\csname PYGdefault@tok@si\endcsname{\let\PYGdefault@bf=\textbf\def\PYGdefault@tc##1{\textcolor[rgb]{0.73,0.40,0.53}{##1}}}
\expandafter\def\csname PYGdefault@tok@s\endcsname{\def\PYGdefault@tc##1{\textcolor[rgb]{0.73,0.13,0.13}{##1}}}
\expandafter\def\csname PYGdefault@tok@na\endcsname{\def\PYGdefault@tc##1{\textcolor[rgb]{0.49,0.56,0.16}{##1}}}
\expandafter\def\csname PYGdefault@tok@cpf\endcsname{\let\PYGdefault@it=\textit\def\PYGdefault@tc##1{\textcolor[rgb]{0.25,0.50,0.50}{##1}}}
\expandafter\def\csname PYGdefault@tok@err\endcsname{\def\PYGdefault@bc##1{\setlength{\fboxsep}{0pt}\fcolorbox[rgb]{1.00,0.00,0.00}{1,1,1}{\strut ##1}}}
\expandafter\def\csname PYGdefault@tok@kp\endcsname{\def\PYGdefault@tc##1{\textcolor[rgb]{0.00,0.50,0.00}{##1}}}
\expandafter\def\csname PYGdefault@tok@no\endcsname{\def\PYGdefault@tc##1{\textcolor[rgb]{0.53,0.00,0.00}{##1}}}
\expandafter\def\csname PYGdefault@tok@nf\endcsname{\def\PYGdefault@tc##1{\textcolor[rgb]{0.00,0.00,1.00}{##1}}}
\expandafter\def\csname PYGdefault@tok@ge\endcsname{\let\PYGdefault@it=\textit}
\expandafter\def\csname PYGdefault@tok@cs\endcsname{\let\PYGdefault@it=\textit\def\PYGdefault@tc##1{\textcolor[rgb]{0.25,0.50,0.50}{##1}}}
\expandafter\def\csname PYGdefault@tok@vc\endcsname{\def\PYGdefault@tc##1{\textcolor[rgb]{0.10,0.09,0.49}{##1}}}
\expandafter\def\csname PYGdefault@tok@sd\endcsname{\let\PYGdefault@it=\textit\def\PYGdefault@tc##1{\textcolor[rgb]{0.73,0.13,0.13}{##1}}}
\expandafter\def\csname PYGdefault@tok@nc\endcsname{\let\PYGdefault@bf=\textbf\def\PYGdefault@tc##1{\textcolor[rgb]{0.00,0.00,1.00}{##1}}}
\expandafter\def\csname PYGdefault@tok@gh\endcsname{\let\PYGdefault@bf=\textbf\def\PYGdefault@tc##1{\textcolor[rgb]{0.00,0.00,0.50}{##1}}}
\expandafter\def\csname PYGdefault@tok@bp\endcsname{\def\PYGdefault@tc##1{\textcolor[rgb]{0.00,0.50,0.00}{##1}}}
\expandafter\def\csname PYGdefault@tok@nn\endcsname{\let\PYGdefault@bf=\textbf\def\PYGdefault@tc##1{\textcolor[rgb]{0.00,0.00,1.00}{##1}}}
\expandafter\def\csname PYGdefault@tok@vi\endcsname{\def\PYGdefault@tc##1{\textcolor[rgb]{0.10,0.09,0.49}{##1}}}
\expandafter\def\csname PYGdefault@tok@se\endcsname{\let\PYGdefault@bf=\textbf\def\PYGdefault@tc##1{\textcolor[rgb]{0.73,0.40,0.13}{##1}}}
\expandafter\def\csname PYGdefault@tok@gs\endcsname{\let\PYGdefault@bf=\textbf}
\expandafter\def\csname PYGdefault@tok@ow\endcsname{\let\PYGdefault@bf=\textbf\def\PYGdefault@tc##1{\textcolor[rgb]{0.67,0.13,1.00}{##1}}}
\expandafter\def\csname PYGdefault@tok@w\endcsname{\def\PYGdefault@tc##1{\textcolor[rgb]{0.73,0.73,0.73}{##1}}}
\expandafter\def\csname PYGdefault@tok@o\endcsname{\def\PYGdefault@tc##1{\textcolor[rgb]{0.40,0.40,0.40}{##1}}}
\expandafter\def\csname PYGdefault@tok@vg\endcsname{\def\PYGdefault@tc##1{\textcolor[rgb]{0.10,0.09,0.49}{##1}}}
\expandafter\def\csname PYGdefault@tok@sr\endcsname{\def\PYGdefault@tc##1{\textcolor[rgb]{0.73,0.40,0.53}{##1}}}
\expandafter\def\csname PYGdefault@tok@m\endcsname{\def\PYGdefault@tc##1{\textcolor[rgb]{0.40,0.40,0.40}{##1}}}
\expandafter\def\csname PYGdefault@tok@il\endcsname{\def\PYGdefault@tc##1{\textcolor[rgb]{0.40,0.40,0.40}{##1}}}
\expandafter\def\csname PYGdefault@tok@sx\endcsname{\def\PYGdefault@tc##1{\textcolor[rgb]{0.00,0.50,0.00}{##1}}}
\expandafter\def\csname PYGdefault@tok@vm\endcsname{\def\PYGdefault@tc##1{\textcolor[rgb]{0.10,0.09,0.49}{##1}}}
\expandafter\def\csname PYGdefault@tok@ch\endcsname{\let\PYGdefault@it=\textit\def\PYGdefault@tc##1{\textcolor[rgb]{0.25,0.50,0.50}{##1}}}
\expandafter\def\csname PYGdefault@tok@nd\endcsname{\def\PYGdefault@tc##1{\textcolor[rgb]{0.67,0.13,1.00}{##1}}}
\expandafter\def\csname PYGdefault@tok@dl\endcsname{\def\PYGdefault@tc##1{\textcolor[rgb]{0.73,0.13,0.13}{##1}}}
\expandafter\def\csname PYGdefault@tok@sh\endcsname{\def\PYGdefault@tc##1{\textcolor[rgb]{0.73,0.13,0.13}{##1}}}
\expandafter\def\csname PYGdefault@tok@gp\endcsname{\let\PYGdefault@bf=\textbf\def\PYGdefault@tc##1{\textcolor[rgb]{0.00,0.00,0.50}{##1}}}
\expandafter\def\csname PYGdefault@tok@ne\endcsname{\let\PYGdefault@bf=\textbf\def\PYGdefault@tc##1{\textcolor[rgb]{0.82,0.25,0.23}{##1}}}
\expandafter\def\csname PYGdefault@tok@mi\endcsname{\def\PYGdefault@tc##1{\textcolor[rgb]{0.40,0.40,0.40}{##1}}}
\expandafter\def\csname PYGdefault@tok@ss\endcsname{\def\PYGdefault@tc##1{\textcolor[rgb]{0.10,0.09,0.49}{##1}}}
\expandafter\def\csname PYGdefault@tok@kt\endcsname{\def\PYGdefault@tc##1{\textcolor[rgb]{0.69,0.00,0.25}{##1}}}
\expandafter\def\csname PYGdefault@tok@fm\endcsname{\def\PYGdefault@tc##1{\textcolor[rgb]{0.00,0.00,1.00}{##1}}}
\expandafter\def\csname PYGdefault@tok@mh\endcsname{\def\PYGdefault@tc##1{\textcolor[rgb]{0.40,0.40,0.40}{##1}}}
\expandafter\def\csname PYGdefault@tok@nb\endcsname{\def\PYGdefault@tc##1{\textcolor[rgb]{0.00,0.50,0.00}{##1}}}
\expandafter\def\csname PYGdefault@tok@kc\endcsname{\let\PYGdefault@bf=\textbf\def\PYGdefault@tc##1{\textcolor[rgb]{0.00,0.50,0.00}{##1}}}
\expandafter\def\csname PYGdefault@tok@gu\endcsname{\let\PYGdefault@bf=\textbf\def\PYGdefault@tc##1{\textcolor[rgb]{0.50,0.00,0.50}{##1}}}
\expandafter\def\csname PYGdefault@tok@nv\endcsname{\def\PYGdefault@tc##1{\textcolor[rgb]{0.10,0.09,0.49}{##1}}}
\expandafter\def\csname PYGdefault@tok@gr\endcsname{\def\PYGdefault@tc##1{\textcolor[rgb]{1.00,0.00,0.00}{##1}}}
\expandafter\def\csname PYGdefault@tok@kr\endcsname{\let\PYGdefault@bf=\textbf\def\PYGdefault@tc##1{\textcolor[rgb]{0.00,0.50,0.00}{##1}}}
\expandafter\def\csname PYGdefault@tok@mf\endcsname{\def\PYGdefault@tc##1{\textcolor[rgb]{0.40,0.40,0.40}{##1}}}
\expandafter\def\csname PYGdefault@tok@cm\endcsname{\let\PYGdefault@it=\textit\def\PYGdefault@tc##1{\textcolor[rgb]{0.25,0.50,0.50}{##1}}}
\expandafter\def\csname PYGdefault@tok@sc\endcsname{\def\PYGdefault@tc##1{\textcolor[rgb]{0.73,0.13,0.13}{##1}}}
\expandafter\def\csname PYGdefault@tok@gi\endcsname{\def\PYGdefault@tc##1{\textcolor[rgb]{0.00,0.63,0.00}{##1}}}
\expandafter\def\csname PYGdefault@tok@cp\endcsname{\def\PYGdefault@tc##1{\textcolor[rgb]{0.74,0.48,0.00}{##1}}}
\expandafter\def\csname PYGdefault@tok@sa\endcsname{\def\PYGdefault@tc##1{\textcolor[rgb]{0.73,0.13,0.13}{##1}}}

\def\PYGdefaultZus{\char`\_}
\def\PYGdefaultZob{\char`\{}
\def\PYGdefaultZcb{\char`\}}

\def\PYGdefaultZam{\char`\&}

\def\PYGdefaultZsh{\char`\#}

\def\PYGdefaultZhy{\char`\-}
\def\PYGdefaultZsq{\char`\'}
\def\PYGdefaultZdq{\char`\"}

\makeatother

\begin{document}

\title{Developing a comprehensive framework for multimodal feature extraction}

\numberofauthors{2}

\author{
\alignauthor Quinten McNamara \\
       \affaddr{University of Texas at Austin}\\
       \affaddr{Austin, Texas}\\
       \email{quinten.mcnamara@utexas.edu}
\alignauthor Alejandro De La Vega \\
       \affaddr{University of Texas at Austin}\\
       \affaddr{Austin, Texas}\\
       \email{delavega@utexas.edu}
\and
\alignauthor Tal Yarkoni\titlenote{Corresponding author.} \\
       \affaddr{University of Texas at Austin}\\
       \affaddr{Austin, Texas}\\
       \email{tyarkoni@utexas.edu}
}

\maketitle

\begin{abstract}
Feature extraction is a critical component of many applied data science workflows. In recent years, rapid advances in artificial intelligence and machine learning have led to an explosion of feature extraction tools and services that allow data scientists to cheaply and effectively annotate their data along a vast array of dimensions---ranging from detecting faces in images to analyzing the sentiment expressed in coherent text. Unfortunately, the proliferation of powerful feature extraction services has been mirrored by a corresponding expansion in the number of distinct interfaces to feature extraction services. In a world where nearly every new service has its own API, documentation, and/or client library, data scientists who need to combine diverse features obtained from multiple sources are often forced to write and maintain ever more elaborate feature extraction pipelines.
To address this challenge, we introduce a new open-source framework for comprehensive multimodal feature extraction. \textit{Pliers} is an open-source Python package that supports standardized annotation of diverse data types (video, images, audio, and text), and is expressly with both ease-of-use and extensibility in mind. Users can apply a wide range of pre-existing feature extraction tools to their data in just a few lines of Python code, and can also easily add their own custom extractors by writing modular classes. A graph-based API enables rapid development of complex feature extraction pipelines that output results in a single, standardized format. We describe the package's architecture, detail its major advantages over previous feature extraction toolboxes, and use a sample application to a large functional MRI dataset to illustrate how pliers can significantly reduce the time and effort required to construct sophisticated feature extraction workflows while increasing code clarity and maintainability.
\end{abstract}

\section{Introduction}

Feature extraction and input annotation are critical elements of many machine learning and data science applications. It is common for data scientists to first extract low-level or high-level semantic features from video, text, or audio input, and subsequently feed these features into a statistical model \cite{hyam_2017, mazloom_rietveld_rudinac_worring_dolen_2016, rangel_cazorla_garcía-varea_martínez-gómez_fromont_sebban_2015}. In many settings, the quality of one's features or annotations can be a greater determinant of an application's success than any subsequent analysis or modeling decisions. For example, the capacity of sentiment analysis models to make sense of speech transcripts extracted from human conversations is inherently limited by the quality of the underlying speech-to-text transcription. If a transcript looks like word salad to human eyes, it is unlikely to be rehabilitated by artificial ones. Fortunately, groundbreaking advances in computer vision, speech recognition, and other domains of artificial intelligence have led to a recent proliferation of highly performant, publicly accessible feature extraction services. Cloud-based APIs developed by major companies such as Google, IBM, and Microsoft, as well as hundreds of smaller startups, now allow data scientists to extract rich annotations from videos, images, audio, and text cheaply and at scale. Consequently, the once monumental challenge of extracting near-human-level feature annotations now appears to be, if not completely overcome, then at least substantially mitigated.

In practice, however, a number of important technical barriers remain. While the proliferation of high-quality feature extraction services has made state-of-the-art machine learning technology widely accessible, harnessing this technology in an optimal way often remains a tedious and confusing process. Which of the hundreds of available speech-to-text, face recognition, or sentiment analysis services or tools should one use? How can one efficiently combine the features extracted using different tools into a cohesive pipeline? And what can one do to ensure that one's codebase remains operational in the face of frequent changes to third-party APIs? There are, at present, no easy answers to such questions; consequently, data scientists developing pipelines that involve extensive feature extraction are often forced to write and maintain highly customized and potentially large codebases. The lack of standardization with respect to feature extraction and data annotation is, in our view, one of the most pervasive and detrimental problems currently facing the applied data science community.

Here we introduce a new feature extraction framework designed to address this critical gap. \textit{Pliers} (available at \\ https://github.com/tyarkoni/pliers) is an open-source Python package that provides a standardized, easy-to-use interface to a wide range of feature extraction tools and services. In contrast to previous feature extraction tools, pliers is inherently multimodal: it supports annotation of diverse data types (video, images, audio, and text), and can seamlessly and dynamically convert between data types as needed (e.g., implicitly transcribing speech in video clips in order to enable application of text-based feature extractors). The package is designed with the twin goals of ease-of-use and extensibility in mind: users can apply a wide range of pre-existing feature extraction tools to their data in just a few lines of Python code, and can also easily add their own custom extractors by writing modular classes. A graph-based API allows users to very rapidly develop complex feature extraction pipelines that output results in a single, standardized format. Thus, pliers can be easily inserted into many existing data science workflows, significantly reducing development time while increasing code clarity and maintainability.

The remaining sections are organized as follows. In section 2, we review relevant previous work, focusing on existing feature extraction toolkits. In section 3, we describe the architecture of the pliers package and highlight several core features with potentially beneficial implications for applied data science workflows. We then demonstrate the flexibility and utility of the package in section 4, where we use pliers to rapidly extract and analyze a broad range of features from the movie stimuli in a large functional neuroimaging dataset. We conclude with a brief discussion of the implications of our work and planned future directions.

\section{Background}
In recent years, major technology companies (e.g., Google and IBM) and smaller startups (e.g., Clarifai and Indico) have released dedicated machine learning services for various computer vision and signal processing tasks. These services can individually be thought of as feature extraction frameworks, each with its own advantages and disadvantages. While these services span an enormous range of features---including everything from face recognition to musical genre detection to part-of-speech tagging---the ability to systematically compare them, or to combine them into cohesive analysis pipelines, is limited by practical considerations. Because each service or tool has its own API, usage requirements, and/or client libraries, there is often considerable overhead involved in accessing and using even a single service, let alone chaining multiple services into integrated pipelines for multimodal feature extraction (e.g., running sentiment analysis on text transcribed from speech audio, in addition to annotating visual sentiment cues).

In an effort to provide common interfaces to different tools, several authors have previously introduced toolboxes that support multiple feature extraction tools. For example, the \textit{SpeechRecognition} package in Python \cite{zhang_2017} supports API calls to several audio transcription services via a single interface; the \textit{MIRToolbox} \cite{lartillot_toiviainen_eerola_2008} supports extraction of both low-level and high-level audio features; the \textit{MMFEAT} feature extraction toolkit \cite{kiela_2016} provides an abstraction for a number of visual and audio feature extraction tools used in natural language processing (NLP) frameworks (e.g., bag of features models, etc.); and the \textit{feature-extraction} package \cite{khosla_bainbridge_torralba_oliva_2013} simplifies extraction of a number of computer vision features commonly used in image classification tasks. Such packages allow their users to call different feature extraction algorithms or services via a standardized interface, reducing development time and avoiding the need to look through each API's documentation for usage examples.

While existing feature extraction packages take important steps towards a unified feature extraction interface, most also have a number of major limitations. First, virtually all previous tools focus on one particular analysis domain---e.g., speech recognition, NLP, or computer vision. We are not aware of any existing toolbox that attempts to cover everything from video to text (though a few support both video and audio---e.g., the feature-extraction package). Second, with a few notable exceptions (e.g., the Essentia audio analysis package \cite{bogdanov_wack_gomez_gulati_herrera_mayor_roma_salamon_zapata_serra_et_al_2013}), most frameworks cannot be easily extended by end users---either because their code is not released under a permissive open-source license, or because the codebase is not designed with modularity in mind. Third, many frameworks focus primarily on traditional low-level visual or audio features use for classification or discrimination tasks, and lack support for high-level perceptual labels of the kind facilitated by recent advances in deep learning. Lastly, and perhaps most importantly, existing frameworks tend to offer a limited degree of abstraction. While many provide a unified interface for querying multiple services or tools, users are still typically responsible for writing supplementary code to pass inputs and handle outputs, chain multiple services, and so on. Our goal in developing the pliers framework was to address each of the above limitations by providing a standardized feature extraction interface that was (i) multimodal, (ii) comprehensive, (iii) extensible, and (iv) easy to use.

\section{Architecture}
\subsection{Overview}
\begin{sloppypar}
With these goals in mind, we sought to develop an intuitive, standardized, and relatively simple API, taking particular cues from the object-oriented transformer interface popularized by the \textit{scikit-learn} machine learning package \cite{buitinck_louppe_blondel_pedregosa_mueller_2013}. At its core, pliers is structured around a hierarchy of \texttt{Transformer} classes that each have a modular, well-defined role in either extracting feature information from an input data object (\texttt{Extractor} classes), or converting the input data object into another data object (\texttt{Converter} classes). As in scikit-learn, the defining feature of a \texttt{Transformer} class in pliers is its implementation of a \texttt{.transform()} method that accepts a single input. The input data types are represented using a separate \texttt{Stim} (short for stimulus) hierarchy that includes subclasses such as \texttt{VideoStim}, \texttt{AudioStim}, and so on. One can thus think of pliers workflows as directed acyclic graphs (DAGs) in which \texttt{Stim} objects are repeatedly transformed as they flow through a graph, culminating in one or more feature extraction steps at the terminal node(s).\end{sloppypar}

The easiest way to illustrate these principles is with a short Python code sample:

\begin{listing}[H]
\begin{Verbatim}[commandchars=\\\{\}, fontsize=\small]
\PYGdefault{n}{video} \PYGdefault{o}{=} \PYGdefault{n}{VideoStim}\PYGdefault{p}{(}\PYGdefault{l+s+s1}{\PYGdefaultZsq{}my\PYGdefaultZus{}video.mp4\PYGdefaultZsq{}}\PYGdefault{p}{)}
\PYGdefault{n}{conv} \PYGdefault{o}{=} \PYGdefault{n}{FrameSamplingConverter}\PYGdefault{p}{(}\PYGdefault{n}{hertz}\PYGdefault{o}{=}\PYGdefault{l+m+mi}{1}\PYGdefault{p}{)}
\PYGdefault{n}{frames} \PYGdefault{o}{=} \PYGdefault{n}{conv}\PYGdefault{o}{.}\PYGdefault{n}{transform}\PYGdefault{p}{(}\PYGdefault{n}{video}\PYGdefault{p}{)}
\PYGdefault{n}{ext} \PYGdefault{o}{=} \PYGdefault{n}{GoogleVisionAPIFaceExtractor}\PYGdefault{p}{()}
\PYGdefault{n}{results} \PYGdefault{o}{=} \PYGdefault{n}{ext}\PYGdefault{o}{.}\PYGdefault{n}{transform}\PYGdefault{p}{(}\PYGdefault{n}{frames}\PYGdefault{p}{)}
\end{Verbatim}

\caption{A simple pliers example}
\label{lst:example}
\end{listing}

This simple snippet of code reads in a video file, converts it into a series of static image frames (sampling frames at a rate of 1 per second), and then uses the Google Cloud Vision API to perform face recognition on each image, returning the aggregated results from all images in a standardized format that can be easily operated on (see \ref{sec:abstraction}). Thus, with very little configuration and code, users are able to perform a complex feature extraction operation using a state-of-the-art face detection API. The benefits of pliers' standardized \texttt{Transformer} API can be further appreciated by considering that we could seamlessly replace the \texttt{GoogleVisionAPIFaceExtractor} call in Listing 1 with one of the many other pliers Extractor classes that support image inputs---for example, a \texttt{ClarifaiAPIExtractor} that performs object recognition using the Clarifai service (clarifai.com), a \texttt{TensorFlowInceptionV3Extractor} that labels images using a pre-trained Tensor Flow inception model \cite{szegedy_vanhoucke_ioffe_shlens_wojna_2016}, and so on. Moreover, as discussed later (\ref{sec:abstraction}), pliers also provides a higher-level graph API that allows users to compactly specify and manage large graphs potentially involving dozens of transformation nodes.

\subsection{Implementation in Python}
Pliers is implemented in the Python programming language, and relies heavily on popular Python scientific computing libraries (most notably, numpy, scipy, and pandas). This choice reflects the widespread usage of Python within the data science community, with many open source libraries available for a variety of tasks. In particular, a large number of feature extraction tools and services are either implemented in Python, or have high-level python bindings or client libraries---greatly simplifying our goal of providing a common interface. The high-level, dynamic nature of the language further facilitates our primary goal of usability and simplicity while remaining relatively performant.
In addition to its core Python dependencies, pliers also depends on the \textit{moviepy} Python package for video and audio file manipulation, which in turn depends on the cross-platform ffmpeg library. Many of the individual feature extractors supported in pliers also have their own dependencies (e.g. the Google Cloud Vision extractors rely on the google-api-python-client library). Pliers is maintained under public version control on GitHub (https://github.com/tyarkoni/pliers). Development follows best practices in the open-source community: we perform continuous integration testing, maintain relatively comprehensive documentation, and include several interactive Jupyter notebooks that exemplify a range of uses of the pliers API. Pliers is platform-independent, though it is tested primarily on Linux and OS X environments.

\subsection{Core Features}
Pliers has a number of core features that distinguish it from previous feature extraction toolboxes, and can help significantly simplify and streamline data scientists' feature extraction pipelines. These include (i) support for a wide range of input modalities; (ii) breadth of feature extractor coverage; (iii) a highly extensible design; and (iv) a high degree of abstraction. We discuss each of these in turn.

\subsubsection{Multimodal support}
Pliers is expressly designed to support multimodal feature extraction from a wide range of input modalities. At present, pliers supports four primary input data types: video, image, audio, and text. Each of these modalities is supported by a hierarchy of classes covering specific use cases (e.g., the base \texttt{ImageStim} class has a \texttt{VideoFrameStim} subclass that represents image frames extracted from video clips). Pliers also implements a generic \texttt{CompoundStim} class that provides slots for any other type of \texttt{Stim} class, allowing users to create custom inputs to feature extractors that require a combination of different input types.  Importantly, and as discussed in more detail in \ref{sec:abstraction}, pliers allows users to seamlessly convert between different \texttt{Stim} classes, making it easy to build complex graphs that integrate feature extractors from multiple modalities. Collectively, the supported input types cover many common use cases in applied data science and scientific research settings (we provide a sample application to functional neuroimaging data in section 4)---and as described in \ref{sec:extensibility}, the \texttt{Stim} hierarchy can be easily extended to support new use cases.

\subsubsection{Breadth}
Pliers seeks to provide unified access to a potentially very large set of feature extraction tools and services. At present, the package supports over 30 different extractors, many of which individually support a large number of more specific models. The current selection of supported extractors has been largely motivated by a proof-of-concept desire to demonstrate the breadth of features that pliers can provide access to; as detailed in the next section, we expect the library of supported tools and services to increase rapidly. Here we describe a partial selection of implemented feature extractors that run the gamut from low-level perceptual analysis to high-level semantic annotations. These include:
\begin{itemize}
\setlength\itemsep{0em}
\item Speech-to-text converters, including wrappers for most of the APIs implemented in the \textit{SpeechRecognition} package \cite{zhang_2017}, in addition to a custom transcription converter that uses IBM's Watson speech-to-text API and has the particular advantage of providing onsets for individual words.
\item A \texttt{TensorFlowInceptionV3Extractor} that uses a pretrained deep convolutional neural network (CNN) model based on the Inception V3 architecture \cite{szegedy2016rethinking} to label objects in images. Pliers seamlessly handles the download, installation, and execution of the associated TensorFlow model and code in the background, illustrating the package's capacity to support not only API-based services, but also a potentially wide range of pretrained, locally-executed models.
\item \begin{sloppypar} An \texttt{OpticalFlowExtractor} that uses OpenCV's Farneback algorithm to quantify the amount of frame-to-frame optical flow in a video. The availability of Python bindings for OpenCV makes it easy to add any number of other OpenCV-based feature extractors (see \ref{sec:extensibility}).\end{sloppypar}
\item Extractors and converters for most of the Google Cloud Vision and Speech API tools---e.g., face detection, object labeling, and speech-to-text conversion. These are some of the most widely used feaure extraction services, and pliers makes it considerably easier to apply them to a diverse range of inputs.
\item \begin{sloppypar} A \texttt{PredefinedDictionaryTextExtractor} that maps individual words onto pre-existing dictionaries accessible via the web. Many of these dictionaries were generated via large-scale psycholinguistic studies involving hundreds of human participants rating thousands of common words; thus, pliers provides easy access to lexical variables ranging from word frequency and contextual diversity to age-of-acquisition norms to affective ratings of word valence.\end{sloppypar}
\item A \texttt{STFTExtractor} that extracts acoustic power in user-configurable frequency bands by applying a short-time Fourier transform to audio input.
\item An \texttt{IndicoAPIExtractor} that interfaces with the Indico API and supports application of a range of models---including sentiment analysis, emotion detection, and personality inference---to text.
\end{itemize}

\subsubsection{Extensibility} \label{sec:extensibility}
The current set of supported feature extractors represents only a fraction of the coverage we intend to eventually provide. We are continuously adding support for new feature extractors and new data types, and encourage others to as well. To date, we have placed emphasis on adding support for (i) high-level APIs capable of providing human-like annotations (e.g., presence of object in an image, sentiment of text), and (ii) other feature extraction toolkits or services that already themselves support multiple extractors (e.g., the Google Cloud Vision APIs, or the Python SpeechRecognition package). However, in principle, virtually any feature extraction tool that operates over one of the supported input data types can be implemented in pliers. To encourage contributions, we have developed pliers as a highly modular, easily-extensible, object-oriented framework. Listing \ref{lst:extractor} provides an example of a simple but fully functional \texttt{Extractor} that takes text inputs and extracts the number of characters in each string. While this example is trivial, it illustrates the speed with which a new \texttt{Extractor} can be developed and immediately embedded into a full pliers workflow. The bare minimum required of a new \texttt{Extractor} class is that it (i) specifies the input data type(s) it operates on (text, video, etc.) and (ii) defines a new \texttt{\_extract()} method that takes a \texttt{Stim} as input, and returns a new object of class \texttt{ExtractorResult} (see \ref{sec:abstraction}).

\begin{listing}[H]
\begin{Verbatim}[commandchars=\\\{\}, fontsize=\small]
\PYGdefault{k}{class} \PYGdefault{n+nc}{LengthExtractor}\PYGdefault{p}{(}\PYGdefault{n}{Extractor}\PYGdefault{p}{):}

    \PYGdefault{n}{\PYGdefaultZus{}input\PYGdefaultZus{}type} \PYGdefault{o}{=} \PYGdefault{n}{TextStim}

    \PYGdefault{k}{def} \PYGdefault{n+nf}{\PYGdefaultZus{}extract}\PYGdefault{p}{(}\PYGdefault{n+nb+bp}{self}\PYGdefault{p}{,} \PYGdefault{n}{stim}\PYGdefault{p}{):}
        \PYGdefault{n}{value} \PYGdefault{o}{=} \PYGdefault{n}{np}\PYGdefault{o}{.}\PYGdefault{n}{array}\PYGdefault{p}{([}\PYGdefault{n+nb}{len}\PYGdefault{p}{(}\PYGdefault{n}{stim}\PYGdefault{o}{.}\PYGdefault{n}{text}\PYGdefault{o}{.}\PYGdefault{n}{strip}\PYGdefault{p}{())]))}
        \PYGdefault{k}{return} \PYGdefault{n}{ExtractorResult}\PYGdefault{p}{(}\PYGdefault{n}{value}\PYGdefault{p}{,} \PYGdefault{n}{stim}\PYGdefault{p}{,} \PYGdefault{n+nb+bp}{self}\PYGdefault{p}{,}
                        \PYGdefault{n}{features}\PYGdefault{o}{=}\PYGdefault{p}{[}\PYGdefault{l+s+s1}{\PYGdefaultZsq{}text\PYGdefaultZus{}length\PYGdefaultZsq{}}\PYGdefault{p}{])}
\end{Verbatim}

\caption{Sample code for an Extractor class that counts the number of words in each text input}
\label{lst:extractor}
\end{listing}

\begin{sloppypar}In many cases, users do not even have to write new extractors in order to add valuable new functionality. Some \texttt{Extractor} classes provide access to named resources that are loaded from a static configuration file, and functionality can thus be extended simply by adding new entries. For example, the \texttt{PredefinedDictionaryExtractor} mentioned above provides access to many web-based dictionary resources that map individual words onto corresponding numerical values (e.g., psycholinguistic databases that provide word frequency, age-of-acquisition, or emotional valence norms for individual words \cite{brysbaert_new_2009, kuperman_stadthagen-gonzalez_brysbaert_2012, warriner_kuperman_brysbaert_2013}). New lookup dictionaries of this kind can be added simply by adding new entries to a JSON configuration file bundled with the pliers package. Listing \ref{lst:predefined} provides a sample (partial) entry from this file that encodes instructions to the \texttt{PredefinedDictionaryExtractor} on how to download and preprocess a popular set of age-of-acquisition norms \cite{kuperman_stadthagen-gonzalez_brysbaert_2012}.\end{sloppypar}

\begin{listing}[H]
\begin{Verbatim}[commandchars=\\\{\}, fontsize=\small]
\PYGdefault{p}{\PYGdefaultZob{}}\PYGdefault{n+nt}{\PYGdefaultZdq{}aoa\PYGdefaultZdq{}}\PYGdefault{p}{:} \PYGdefault{p}{\PYGdefaultZob{}}
    \PYGdefault{n+nt}{\PYGdefaultZdq{}title\PYGdefaultZdq{}}\PYGdefault{p}{:} \PYGdefault{l+s+s2}{\PYGdefaultZdq{}Age\PYGdefaultZhy{}of\PYGdefaultZhy{}acquisition (AoA) norms for}
\PYGdefault{l+s+s2}{    over 50 thousand English words\PYGdefaultZdq{}}\PYGdefault{p}{,}
    \PYGdefault{n+nt}{\PYGdefaultZdq{}description\PYGdefaultZus{}url\PYGdefaultZdq{}}\PYGdefault{p}{:} \PYGdefault{l+s+s2}{\PYGdefaultZdq{}http://crr.ugent.be/archives/806\PYGdefaultZdq{}}\PYGdefault{p}{,}
    \PYGdefault{n+nt}{\PYGdefaultZdq{}source\PYGdefaultZdq{}}\PYGdefault{p}{:} \PYGdefault{l+s+s2}{\PYGdefaultZdq{}Kuperman, V., Stadthagen\PYGdefaultZhy{}Gonzalez, H.,}
\PYGdefault{l+s+s2}{    \PYGdefaultZam{} Brysbaert, M. (2012). Age\PYGdefaultZhy{}of\PYGdefaultZhy{}acquisition ratings}
\PYGdefault{l+s+s2}{    for 30,000 English words. Behavior Research Methods,}
\PYGdefault{l+s+s2}{    44(4), 978\PYGdefaultZhy{}990.\PYGdefaultZdq{}}\PYGdefault{p}{,}
    \PYGdefault{n+nt}{\PYGdefaultZdq{}url\PYGdefaultZdq{}}\PYGdefault{p}{:} \PYGdefault{l+s+s2}{\PYGdefaultZdq{}http://crr.ugent.be/papers/AoA\PYGdefaultZus{}51715\PYGdefaultZus{}words.zip\PYGdefaultZdq{}}\PYGdefault{p}{,}
    \PYGdefault{n+nt}{\PYGdefaultZdq{}format\PYGdefaultZdq{}}\PYGdefault{p}{:} \PYGdefault{l+s+s2}{\PYGdefaultZdq{}xls\PYGdefaultZdq{}}\PYGdefault{p}{,}
    \PYGdefault{n+nt}{\PYGdefaultZdq{}language\PYGdefaultZdq{}}\PYGdefault{p}{:} \PYGdefault{l+s+s2}{\PYGdefaultZdq{}english\PYGdefaultZdq{}}\PYGdefault{p}{,}
    \PYGdefault{n+nt}{\PYGdefaultZdq{}index\PYGdefaultZdq{}}\PYGdefault{p}{:} \PYGdefault{l+s+s2}{\PYGdefaultZdq{}Word\PYGdefaultZdq{}}
\PYGdefault{p}{\PYGdefaultZcb{}\PYGdefaultZcb{}}
\end{Verbatim}

\caption{Sample resource available through the PredefinedDictionaryExtractor}
\label{lst:predefined}
\end{listing}

In still other cases, harnessing new functionality is even easier than adding new entries to a config file. Because many of the feature extraction APIs that pliers supports allow users to easily control the remote resource or model used in feature extraction, specifying a different resource can be as easy as passing in the appropriate parameter. For example, the \texttt{IndicoAPIExtractor}---which provides access to the Indico.io service's text analysis tools---accepts an initialization argument that specifies which of its available models (e.g., `sentiment', `emotion', or `personality') to use. As new models are introduced to the Indico.io service, they automatically become available to pliers users.

Extensibility in pliers is not limited to adding new transformers; new input data types can also be readily added by creating new \texttt{Stim} classes---most commonly by simply subclassing an existing \texttt{Stim} class. For example, while pliers currently does not provide a built-in interface to social media services such as Facebook or Twitter, our near-term roadmap includes plans to develop tools that make it easy to retrieve tweets and Facebook posts as native pliers objects (e.g., \texttt{TweetStim} or \texttt{FacebookPostStim} classes that contain metadata in addition to multimedia)---at which point users will be immediately and effortlessly able to apply any of the compatible feature extractors available in pliers.

\subsubsection{Abstraction} \label{sec:abstraction}
One of our primary goals in developing pliers is to to create a framework that makes feature extraction as easy as possible, enabling data scientists to develop complex feature extraction pipelines using clearer, more compact code. To this end, pliers features an extremely high level of abstraction. Many steps that would require explicit routines in other feature extraction packages are performed implicitly in pliers. Here we highlight four particular features in pliers that greatly simplify the feature extraction process in comparison to other feature extraction packages or bespoke pipelines.

{\bf Implicit \texttt{Stim} conversion.} Pliers is capable of implicitly converting between different data types to facilitate feature extraction on an ad hoc basis. For example, suppose one wishes to apply a sentiment extractor to the dialogue in a series of videos. Accomplishing this requires a user to first extract the audio track from the video, and then apply speech-to-text transcription to the audio. Helpfully, pliers is usually able to identify and implicitly execute conversions between different input types, minimizing the user's workload and considerably streamlining the feature extraction pipeline. For example, the two code snippets in Listing \ref{lst:implicit} produce identical results.

\begin{listing}[H]
\begin{Verbatim}[commandchars=\\\{\}, fontsize=\small]
\PYGdefault{c+c1}{\PYGdefaultZsh{} Option A: explicit conversion}
\PYGdefault{n}{conv1} \PYGdefault{o}{=} \PYGdefault{n}{VideoToAudioConverter}\PYGdefault{p}{()}
\PYGdefault{n}{audio} \PYGdefault{o}{=} \PYGdefault{n}{conv1}\PYGdefault{o}{.}\PYGdefault{n}{transform}\PYGdefault{p}{(}\PYGdefault{n}{video}\PYGdefault{p}{)}
\PYGdefault{n}{conv2} \PYGdefault{o}{=} \PYGdefault{n}{AudioToTextConverter}\PYGdefault{p}{()}
\PYGdefault{n}{text} \PYGdefault{o}{=} \PYGdefault{n}{conv2}\PYGdefault{o}{.}\PYGdefault{n}{transform}\PYGdefault{p}{(}\PYGdefault{n}{audio}\PYGdefault{p}{)}
\PYGdefault{n}{ext} \PYGdefault{o}{=} \PYGdefault{n}{IndicoAPIExtractor}\PYGdefault{p}{(}\PYGdefault{n}{model}\PYGdefault{o}{=}\PYGdefault{l+s+s1}{\PYGdefaultZsq{}sentiment\PYGdefaultZsq{}}\PYGdefault{p}{)}
\PYGdefault{n}{result} \PYGdefault{o}{=} \PYGdefault{n}{ext}\PYGdefault{o}{.}\PYGdefault{n}{transform}\PYGdefault{p}{(}\PYGdefault{n}{text}\PYGdefault{p}{)}

\PYGdefault{c+c1}{\PYGdefaultZsh{} Option B: implicit conversion}
\PYGdefault{n}{ext} \PYGdefault{o}{=} \PYGdefault{n}{IndicoAPIExtractor}\PYGdefault{p}{(}\PYGdefault{n}{model}\PYGdefault{o}{=}\PYGdefault{l+s+s1}{\PYGdefaultZsq{}sentiment\PYGdefaultZsq{}}\PYGdefault{p}{)}
\PYGdefault{n}{result} \PYGdefault{o}{=} \PYGdefault{n}{ext}\PYGdefault{o}{.}\PYGdefault{n}{transform}\PYGdefault{p}{(}\PYGdefault{n}{video}\PYGdefault{p}{)}
\end{Verbatim}
\caption{Explicit vs. implicit stimulus conversion}
\label{lst:implicit}
\end{listing}

Here, pliers implicitly identifies and applies a valid conversion trajectory that transforms a video into text (by stripping the audio track from the video and submitting it to a speech-to-text API). Pliers provides package-level control over which \texttt{Converter} to use in cases of ambiguity (e.g., in the above example, the user could specify which of several speech-to-text services they prefer to work with). Moreover, all \texttt{Stim} instances retain a comprehensive internal log of applied transformations, allowing the user to easily determine what steps were taken in order to produce the final result.

\begin{sloppypar}{\bf Native handling of iterable inputs.} Every pliers \texttt{Transformer} (including all \texttt{Extractor}s and \texttt{Converter}s) is inherently iterable-aware, and can be passed an iterable (i.e., a Python list, tuple, or generator) of \texttt{Stim} objects rather than just a single \texttt{Stim}. The transformation will then be applied independently to each \texttt{Stim}. Furthermore, some \texttt{Stim} classes are naturally iterable. For example, a \texttt{VideoStim} is made up of a series of \texttt{VideoFrameStim}s, and a \texttt{ComplexTextStim} is made up of \texttt{TextStim}s. For efficiency and memory conservation reasons, the elements will typically be retrieved using Python generator expressions (functions that support lazy evaluation of iterable objects) rather than lists.
\end{sloppypar}

\begin{figure}
\includegraphics[width=9cm]{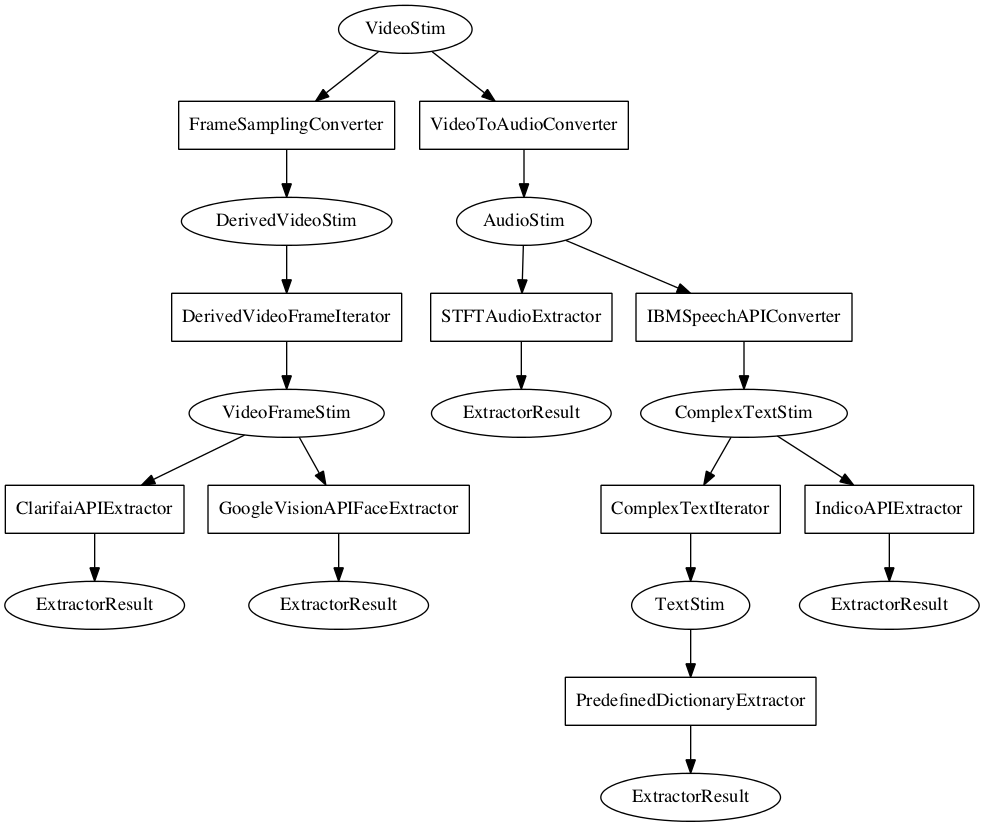}
\caption{Diagram of the full graph generated by the code in Listing \ref{lst:graph}.}
\label{fig:graph}
\end{figure}

{\bf Graph API.} Pliers also implements a unique graph interface that enables users to effortlessly go from complex stimuli to a potentially large number of features extracted from multiple modalities. The interface allows users to specify \texttt{Converter} and \texttt{Extractor} nodes through which a stimulus can be run. Suppose we have a video that we want to tag for visual annotation, audio features, lexical variables (e.g., the normative frequency of each spoken word), and spoken word sentiment. This seemingly complex task can be simply configured using the code in Listing \ref{lst:graph}. Moreover, pliers allows users to easily visualize the executed graph by leveraging the graphviz package; Figure \ref{fig:graph} displays the graph generated by the code in Listing \ref{lst:graph}. Note that, as discussed above, it is usually not necessary to explicitly specify \texttt{Stim} conversion steps, as these will be detected and injected implicitly (and hence the diagram in Figure \ref{fig:graph} contains many more nodes than were specified in Listing \ref{lst:graph}).

\begin{listing}[H]
\begin{Verbatim}[commandchars=\\\{\}]
\PYGdefault{n}{clips} \PYGdefault{o}{=} \PYGdefault{p}{[}\PYGdefault{l+s+s1}{\PYGdefaultZsq{}video1.mp4\PYGdefaultZsq{}}\PYGdefault{p}{,} \PYGdefault{l+s+s1}{\PYGdefaultZsq{}video2.mp4\PYGdefaultZsq{}}\PYGdefault{p}{]}
\PYGdefault{n}{g} \PYGdefault{o}{=} \PYGdefault{n}{Graph}\PYGdefault{p}{([}
    \PYGdefault{p}{(}\PYGdefault{n}{FrameSamplingConverter}\PYGdefault{p}{(}\PYGdefault{n}{hertz}\PYGdefault{o}{=}\PYGdefault{l+m+mi}{1}\PYGdefault{p}{),}
     \PYGdefault{p}{[}\PYGdefault{l+s+s1}{\PYGdefaultZsq{}ClarifaiAPIExtractor\PYGdefaultZsq{}}\PYGdefault{p}{,}
      \PYGdefault{l+s+s1}{\PYGdefaultZsq{}GoogleVisionAPIFaceExtractor\PYGdefaultZsq{}}\PYGdefault{p}{]),}
    \PYGdefault{n}{STFTAudioExtractor}\PYGdefault{p}{(}\PYGdefault{n}{hop\PYGdefaultZus{}size}\PYGdefault{o}{=}\PYGdefault{l+m+mi}{1}\PYGdefault{p}{,} \PYGdefault{n}{freq\PYGdefaultZus{}bins}\PYGdefault{o}{=}\PYGdefault{l+m+mi}{5}\PYGdefault{p}{),}
    \PYGdefault{n}{PredefinedDictionaryExtractor}\PYGdefault{p}{(}
      \PYGdefault{p}{[}\PYGdefault{l+s+s1}{\PYGdefaultZsq{}SUBTLEXusfrequencyabove1/Lg10WF\PYGdefaultZsq{}}\PYGdefault{p}{,}
       \PYGdefault{l+s+s1}{\PYGdefaultZsq{}concreteness/Conc.M\PYGdefaultZsq{}}\PYGdefault{p}{]),}
    \PYGdefault{n}{IndicoAPIExtractor}\PYGdefault{p}{(}\PYGdefault{n}{models}\PYGdefault{o}{=}\PYGdefault{p}{[}\PYGdefault{l+s+s1}{\PYGdefaultZsq{}sentiment\PYGdefaultZsq{}}\PYGdefault{p}{])}
\PYGdefault{p}{])}
\PYGdefault{n}{results} \PYGdefault{o}{=} \PYGdefault{n}{g}\PYGdefault{o}{.}\PYGdefault{n}{run}\PYGdefault{p}{(}\PYGdefault{n}{clips}\PYGdefault{p}{)}
\end{Verbatim}

\caption{Sample graph generation code}
\label{lst:graph}
\end{listing}

\begin{sloppypar}{\bf Consolidated output.} One of the major sources of busywork data scientists often face when building feature extraction pipelines is the need to postprocess and reformat the results returned by different feature extraction tools. Pliers addresses this problem by ensuring that all \texttt{Extractor} classes output an \texttt{ExtractorResult} object. This is a lightweight container that represents all of the extracted feature information returned by the \texttt{Extractor} in a common format, and also stores references to the \texttt{Stim} and \texttt{Extractor} objects used to generate the result. The raw extracted feature values are stored in the \texttt{.data} property, but typically, users will want to work with the data in a more convenient format. Fortunately, every \texttt{ExtractorResult} instance exposes a \texttt{.to\_df()} method that returns a formatted pandas DataFrame (a tabular data type widely used in Python data science applications). Additionally, pliers provides built-in tools to easily merge the data from multiple \texttt{ExtractorResult} instances; by default, executing a Graph such as the one in Listing \ref{lst:graph} will return a single merged pandas DataFrame that contains all extracted feature information for all processed inputs, along with added timestamp and duration columns that facilitate further manipulation, plotting, and analysis of the results.\end{sloppypar}

\begin{figure*}[t]
\centering
\includegraphics[width=18cm]{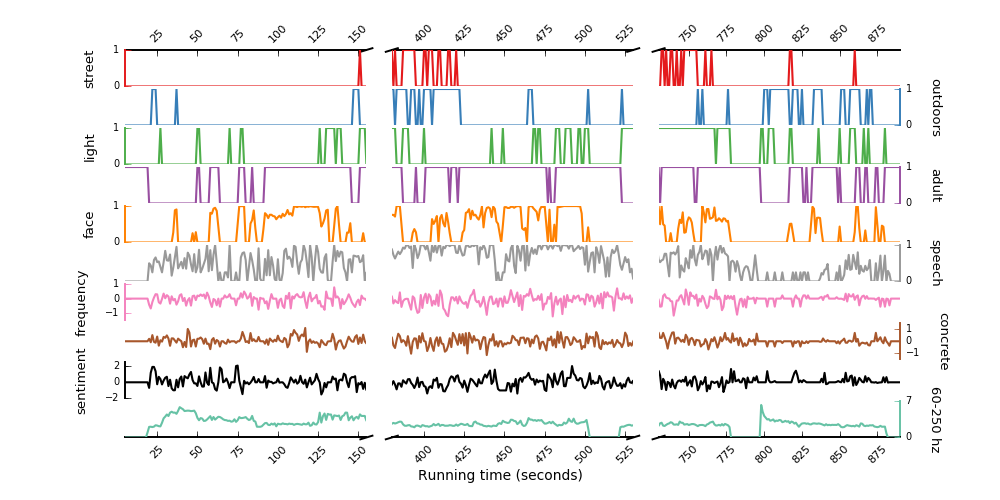}
\caption{Timeline of automatically extracted features across three periods of the movie stimulus. We used the \texttt{ClarifaiAPIExtractor} to label four image classes (`street', `outdoors'€™, `light'€™ and `adult'€™) and the \texttt{GoogleFaceAPIExtractor} to determine the probability that a face was present. Next, we applied the \texttt{IBMSpeechAPIConverter} to transcribe the movie audio and detect the presence of speech (`speech'). For each transcribed word we used a \texttt{PredefinedDictionaryExtractor} to extract lexical norms for word `frequency' and `concreteness', and used the \texttt{IndicoAPIExtractor} to quantify `sentiment'. Finally, a \texttt{STFTAudioExtractor} was used to quantify acoustic power within the 60-250 hz frequency range (where most human speech is expressed).}
\label{fig:timeline}
\end{figure*}

\section{Application to functional MRI}

\begin{figure*}[t]
\centering
\includegraphics[width=18cm]{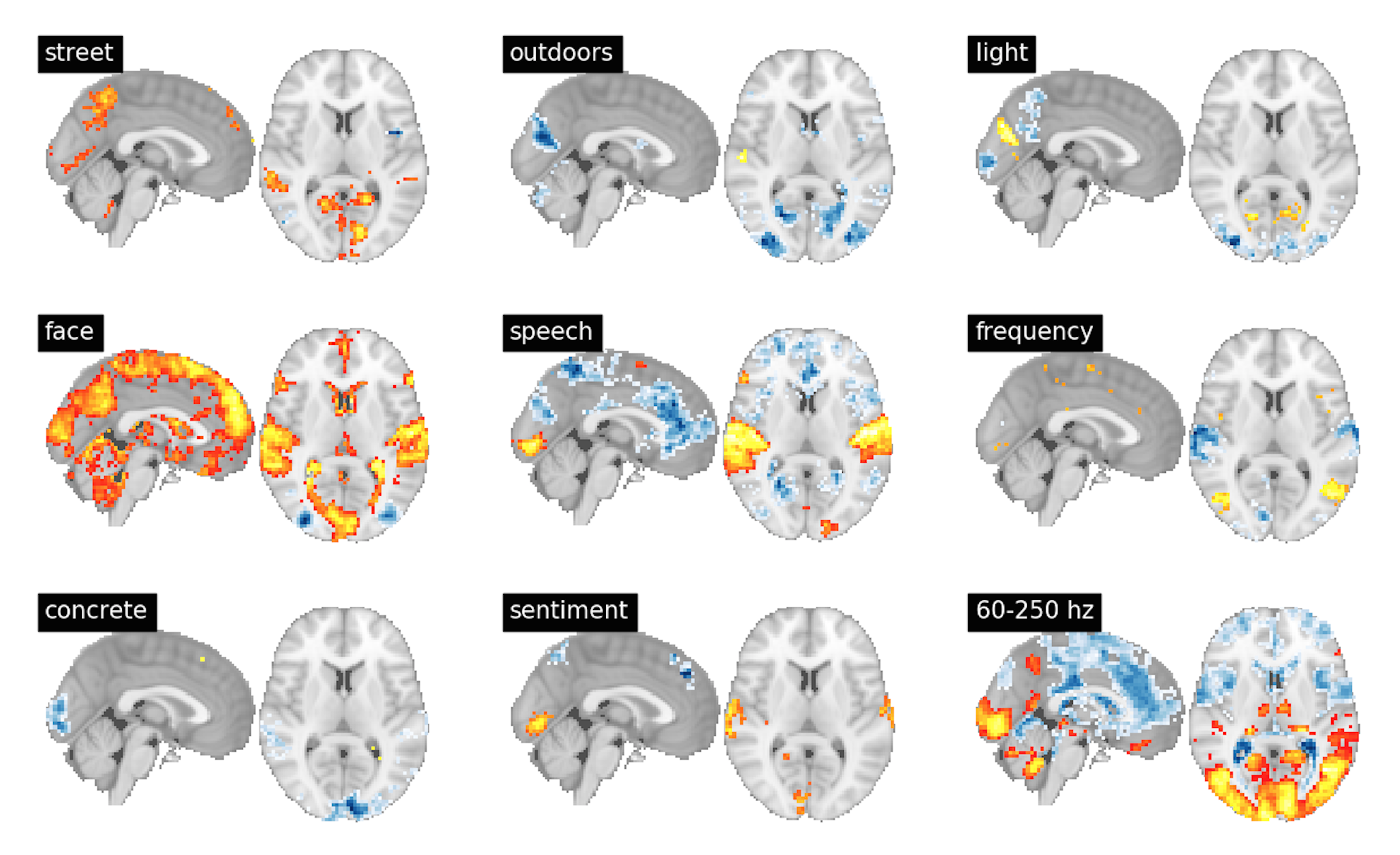}
\caption{Pattern of brain activity associated with automatically extracted features from an audio visual movie. Labels (e.g., ``street'', ``frequency'', etc.) correspond to the rows and descriptions in Fig. \ref{fig:timeline}. Yellow areas indicate voxels that were positively associated with the predictor, whereas blue areas indicated voxels with a negative association. More intense colors reflect stronger effects. Maps were thresholded at a statistical threshold of p $<$ .001 for display purposes.}
\label{fig:brains}
\end{figure*}

To demonstrate the utility of having a standardized, high-level, multimodal feature extraction framework, we applied pliers to the domain of human functional neuroimaging. A primary goal of neuroimaging research is to understand how the brain processes and integrates information from different sensory modalities. In pursuit of this goal, researchers traditionally employ ``factorial" designs that experimentally contrast the brain's response to different sets of stimuli which vary systematically along one or more dimensions. For example, to study the neural substrates of face processing, one might compare the brain activity elicited by images of faces to the responses elicited by other image categories such as buildings, animals, or outdoor scenes. Critically, if both sets of images do not differ along irrelevant dimensions---such as luminosity, size or color---differences in brain activity should reflect face-specific processing \cite{kanwisher_yovel_2006}.

Unfortunately, this idealized strategy has several important limitations. Most notably, in practice, it is rarely possible to ensure that the stimuli in experimental conditions vary \textit{only} along the target dimension of interest. Almost invarably, stimuli vary on many additional dimensions that are not of interest, and may confound the neural response, but are omitted from the statistical model. Moreover, by studying the phenomena of interest under highly controlled, simplifed conditions, experiments often lose their ecological validity---that is, they poorly approximate the complexities of real world scene processing and understanding.

An alternative approach is to use rich, ``naturalistic" stimuli, such as audiovisual movies, to examine the dynamic neural response to a large number of potential features. By maximizing the number of predictors in one's statistical model, one can better control for potential confounds, while simultaneously maximizing ecological validity. However, the high cost of manually annotating a large number of features has made such naturalistic approaches relatively rare. Fortunately, pliers allows us to generate large sets of potential predictors rapidly and automatically.

Here, we modeled the brain's responses to a variety of automatically extracted features using publicly accessible data from the seminal Human Connectome Project (HCP \cite{glasser_sotiropoulos_wilson_coalson_fischl_andersson_xu_jbabdi_webster_polimeni_et_al_2013, van_essen_smith_barch_behrens_yacoub_ugurbil_2013}). In this study, participants underwent functional magnetic resonance imaging (fMRI) while watching a series of short films, for a total of 60 minutes of audio-visual stimulation. We applied the pliers Graph from Listing \ref{lst:graph} to the video stimuli used in the study, resulting in a timeline of feature vectors that span the full duration of the scans (an extract is displayed in Figure \ref{fig:timeline}). We then used the Nipype workflow engine \cite{Gorgolewski2011}---which provides uniform interfaces to neuroimaging analysis packages such as FSL \cite{smith_jenkinson_woolrich_beckmann_behrens_johansen-berg_bannister_luca_drobnjak_flitney_2004}---to predict brain activity using a subset of 10 automatically extracted features across 35 subjects.

We omit the full details of the fMRI analysis pipeline and statistical model here, as our goal here is to illustrate the practical application of pliers in complex feature extraction workflows rather than to draw conclusions about the neural correlates of different features. The key point is simply that we model the neural response (technically, the hemodynamic response, which is an indirect measure of neuronal activity) as a linear function of the extracted features. An abridged version of the mixed model we fit is:

$$ Y_{it} = \beta_0 + \beta_1X_{1it} + ... + \beta_kX_{kit} + u_{0i} + u_{1i}X_{1it} + ... + u{ki}X_{kit} + e_{it}, $$

where $Y_{it}$ is the $i^{th}$ participant's neural response at time $t$, $\beta_0$ is a fixed intercept, $\beta_k$ is the estimated fixed regression coefficient for the $k^{th}$ extracted feature, and $X_{kit}$ is the value of the $k^{th}$ extracted feature at time $t$ in the $i^{th}$ participant, as obtained using pliers. The $u$ terms represents random subject intercepts and slopes intended to account for between-subject variance, and $e_{it}$ is the residual model error. For the sake of brevity, we omit a number of other terms that were included in the actual statistical model in order to account for fMRI-specific modeling concerns (e.g., temporal autocorrelation, low-frequency noise, participant head movement, etc.).

This standard neuroimaging model is fit separately in each of over 200,000 brain voxels, resulting in whole-brain activation maps displaying those voxels across the brain that were consistently associated (as assessed using a standard significance test, and correcting for multiple comparisons) with each of the extracted features. We were able to identify distributed patterns of neural activity that were statistically significantly associated with all but one of the extracted features (Figure \ref{fig:brains}). While interpretation of these results is of little importance here, we note in passing that a number of the patterns exhibited in these statistical maps replicated the neural correlates previously identified in more conventional factorial experimental designs. For example, in the auditory domain, speech was associated with activity in areas established language processing regions, such as the superior frontal gyrus (SFG) \cite{scott_johnsrude_2003}, and in the visual domain, distinct image tags such as ``outdoors", ``street" and ``light", were associated with differential activation in brain regions important for natural scene recognition, such as visual and retrosplenial cortices \cite{walther_caddigan_fei-fei_beck_2009}.

In the present case study, we demonstrated how automated feature extraction can be used to rapidly test novel hypothesis in naturalistic fMRI experiments. Pliers allows researchers to quickly and flexibly test a potentially large number of neuroscientifically interesting hypotheses using pre-existing fMRI datasets. Furthermore, the rapid and automated nature allows researchers to quickly generate results and more easily replicate their analyses in new datasets, increasing the generalizability of their findings. Although more work is necessary to optimize statistical modeling in naturalistic datasets---in part due to concerns of collinearity between extracted features---the present approach enables a new class of neuroimaging study that promises to maximize ecological validity and encourage re-use of openly accessible datasets. Of course, this example highlights just one among many potential uses for our feature extraction framework.

\section{Limitations}
In its present state, pliers is already fully functional and suitable for use in feature extraction pipelines across a range of applied data science and research settings. Nevertheless, pliers remains under active development, and currently has a number of limitations worth noting. First, while pliers already supports a number of state-of-the-art, widely used feature extraction tools and services, there are hundreds, if not thousands, of others that are currently unsupported but that could be readily added to the package. The utility of pliers should continue to increase as we and others continue to add new extractors to the package.

Second, our emphasis on ease-of-use and a high degree of abstraction sometimes comes at the cost of decreased performance or a loss of fine-grained control. While pliers supports basic caching of transformer outputs in order to prevent repeated application of computationally (or, in the case of some APIs, financially) expensive transformations, and adopts lazy evaluation patterns whenever possible, it otherwise makes no explicit effort to optimize memory consumption or CPU usage. Similarly, many of the supported Extractors do not provide tight control over every parameter of the underlying tool or algorithm. For example, the \texttt{OpticalFlowExtractor} (which quantifies the amount of motion between consecutive movie frames) offers no control over the parameters of the Farneback algorithm it relies on. Instead, pliers places emphasis on easy specification of those parameters most likely to vary across common use cases. For example, the \texttt{ClarifaiAPIExtractor} allows the user to specify (upon initialization) the Clarifai model to query, as well as any specific classes or labels to retrieve (e.g., to quantify the likelihood that an "animal" is present in each image).

Lastly, an important potential challenge for pliers is the continued maintenance of the package in light of not-infrequent version upgrades to the various tools and services that pliers supports. We expect that the third-party APIs pliers leverages for most of its extractors will improve over time, which means that some extractor code will occasionally need to be reconfigured to account for external API changes. To combat this, we intend to diligently keep up-to-date with the supported APIs and services. Maintenance of the pliers codebase is facilitated by the use of a continuous integration testing service (Travis-CI) that notify the developers whenever a previously functional transformer begins to fail.

In addition to addressing the above limitations, there are a number of major enhancements on our near-term development agenda. These include basic parallelization functionality, integration with social media APIs, and persisting intermediate conversion results to a database or disk.

\section{Conclusion}
We have outlined the key properties and major benefits of a new framework for high-level, standardized, multimodal feature extraction. Pliers enables users to easily and quickly extract a variety of image, audio, and text features using a standardized, highly extensible, and relatively simple interface, providing an effective solution to one of the most common problems faced by applied data scientists.

\section{Acknowledgments}
This work was supported by a National Institutes of Health award to TY (NIH award R01MH109682).


\bibliographystyle{abbrv}
\bibliography{ref}

\end{document}